\documentclass[lettersize,journal]{IEEEtran}
\IEEEoverridecommandlockouts
\usepackage{cite}
\usepackage{amsmath,amssymb,amsfonts}
\usepackage{algorithmic}
\usepackage{graphicx}
\usepackage{textcomp}
\usepackage{xcolor}
\usepackage{caption} % for subfigures
\usepackage{subcaption} % for subfigures
\usepackage{booktabs}
\usepackage[hidelinks]{hyperref}

\usepackage{algorithm}
\usepackage{array}
\usepackage{textcomp}
\usepackage{stfloats}
\usepackage{url}
\usepackage{verbatim}
\usepackage{cite}
\usepackage{orcidlink}
\usepackage{fancyhdr}
\makeatletter
\let\old@headrule\headrule
\renewcommand{\headrule}{\if@fancyplain\let\headrulewidth\plainheadrulewidth\fi\old@headrule}
\makeatother

\renewcommand{\headrulewidth}{0pt}

\makeatletter
\def\ps@IEEEtitlepagestyle{%
  \def\@oddhead{\mbox{}\scriptsize\rightmark \hfil}%
  \def\@evenhead{\scriptsize\thepage \hfil \leftmark\mbox{}}%
  \def\@oddfoot{\hfil \mbox{}\parbox{6.5in}{\centering
  \footnotesize \textcopyright 2024 IEEE. Personal use of this material is permitted.
  Permission from IEEE must be obtained for all other uses, in any current or future
  media, including reprinting/republishing this material for advertising or promotional
  purposes, creating new collective works, for resale or redistribution to servers or
  lists, or reuse of any copyrighted component of this work in other works.}\hfil \mbox{}}%
  \def\@evenfoot{\mbox{}\parbox{5.5in}{\centering}\hfil \mbox{}\hfil}%
}

\makeatother
\IEEEoverridecommandlockouts

\begin{document}

\title{
%Experience-Driven Balancing of competitive Game Levels with Reinforcement Learning\\
Simulation-Driven Balancing of Competitive Game Levels with Reinforcement Learning\\

\thanks{This research was supported by the Volkswagen Foundation (Project: Consequences of Artificial Intelligence on Urban Societies, Grant 98555).}
}

%\markboth{Transaction on Games ~Vol.~X, No.~X, Month~Year}

\author{Florian Rupp \orcidlink{0000-0001-5250-8613}, Manuel Eberhardinger \orcidlink{0009-0009-2897-9250}, Kai Eckert \orcidlink{0000-0002-5423-561X}
}

\maketitle

\IEEEpubidadjcol

\begin{abstract}
The balancing process for game levels in competitive two-player contexts involves a lot of manual work and testing, particularly for non-symmetrical game levels.
In this work, we frame game balancing as a procedural content generation task and propose an architecture for automatically balancing of tile-based levels within the PCGRL framework (procedural content generation via reinforcement learning).
Our architecture is divided into three parts: (1) a level generator, (2) a balancing agent, and (3) a reward modeling simulation. Through repeated simulations, the balancing agent receives rewards for adjusting the level towards a given balancing objective, such as equal win rates for all players.
To this end, we propose new swap-based representations to improve the robustness of playability, thereby enabling agents to balance game levels more effectively and quickly compared to traditional PCGRL. By analyzing the agent's swapping behavior, we can infer which tile types have the most impact on the balance. We validate our approach in the Neural MMO (NMMO) environment in a competitive two-player scenario.
In this extended conference paper, we present improved results, explore the applicability of the method to various forms of balancing beyond equal balancing, compare the performance to another search-based approach, and discuss the application of existing fairness metrics to game balancing.
\end{abstract}


%\captionsetup[sub]{font=normalfont}
\section{Introduction}
 
Level design is a key concept when creating games. In order to keep players engaged, a balance must be struck between a challenging and enjoyable experience. This is generally not an easy task, as it also depends on the skill and experience of the players~\cite{schreiber_game_2021}.
In addition, game levels for competitive multiplayer games must be designed to be balanced towards equal initial win chances for all players. Imbalanced games will lead to boredom or frustration, and players will quit playing~\cite{andrade_dynamic_2006,becker_what_2020}.

To ensure balance through level design, game designers often rely on nearly (point) symmetric map architectures. This can be seen in popular competitive e-sports titles such as League of Legends, DotA~2 or Starcraft~2, but also in other competitive tile-based games such as Advance Wars or Bomberman.
In addition to symmetric levels, alternative approaches are also possible. Non-symmetric levels offer more variety and can create new ways for playful creativity to be entertaining and challenging.

We formulate the level balancing task as a procedural content generation (PCG) task. In previous works, several approaches have been proposed using (PCG) for level balancing, such as search-based approaches~\cite{togelius_search-based_2011,sorochan_generating_2022}, evolutionary algorithms~\cite{morosan_automated_2017,lanzi_evolving_2014,lara-cabrera_balance_2014,karavolos_using_2018,mesentier_silva_evolving_2019} or graph grammars~\cite{kowalski_strategic_2018}.
In this work, we introduce the use of reinforcement learning (RL) to balance tile-based levels. Once a RL model is trained, it can generate levels fast and is less dependent on randomness compared to evolutionary methods. We apply RL for balancing by designing a reward function based on multiple simulation runs using scripted agents. While this approach becomes computationally intensive due to the simulations for each reward calculation, the advantage of RL over other search-based approaches is that it learns during training. This reduces the need for unnecessary simulations in the inference step, unlike e.g., evolutionary algorithms, which start over each time.
In addition, we design our method to be domain independent, unlike graph grammars or search-based approaches.

\begin{figure}
     \centering
        \includegraphics[width=2.8in]{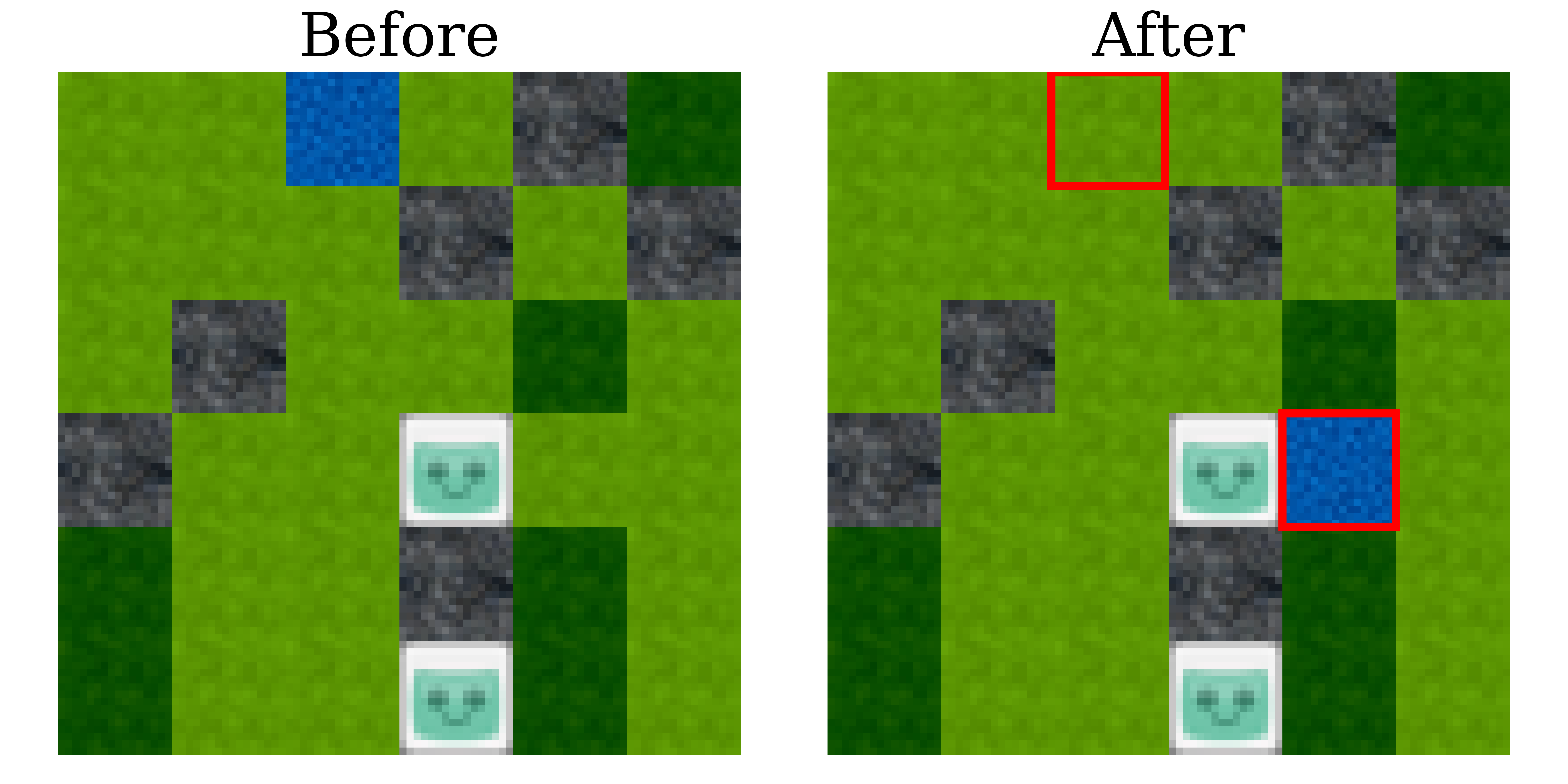}
    \caption{
In our game environment, two players must forage for resources like food (dark green) and water (blue) to survive longest. By swapping the highlighted tiles (red), the trained model achieved a more balanced game in simulated game runs. 
}
     \label{fig:gen-level1}
\end{figure}

Therefore, we use the Procedural Content Generation via Reinforcement Learning framework (PCGRL)~\cite{khalifa_pcgrl_2020} which has recently been introduced for the generation of tile-based levels. Together with the newly introduced swapping representations, our method can be viewed as a fine-tuning task to optimize a generated level towards a certain balance.
Our method is evaluated on the open-source Neural Massively Multiplayer Online (NMMO) environment~\cite{suarez_neural_2019} which was designed for competitive multiplayer research. We frame the problem as a resource gathering and survival game. Figure~\ref{fig:gen-level1} shows an example of how our method balanced a generated level (left) by swapping highlighted tiles (right). The code used in this research is available on Github\footnote{https://github.com/FlorianRupp/pcgrl-simulation-driven-balancing}.

Our contributions are:
\begin{itemize}
    \item A domain-independent architecture for learning to automatically alter game levels towards a specified balance with RL. 
    \item A novel swap-based representation pattern to frame the problem as a Markov decision process.
    \item An experimental study to evaluate the proposed architecture against the original PCGRL method.
\end{itemize}

This paper is an extended version of the conference paper~\cite{rupp_balancing_2023}. We provide improved results, a detailed discussion of how existing fairness metrics can be transferred to game balancing, a comparison with a search-based approach, and a study of how this can be used for any balancing and not just equal balancing. Therefore, we revise the reward function and explain its derivation from the statistical parity metric~\cite{dwork_fairness_2012}.

The structure of this paper is as follows: We give a brief overview of related work~(\ref{sec:related-work}) and the background~(\ref{sec:background}). In Section~\ref{sec:method} the method is described and the implementation details in~\ref{sec:details} subsequently. Experiments and results are presented in~\ref{sec:experiments}, followed by an extended study on the generalized approach of an arbitrary balancing in~\ref{sec:extensions}. The discussion is in Section~\ref{sec:discussion}, the conclusion in~\ref{sec:conclusion}.

\section{Related Work}
\label{sec:related-work}
% game balancing
Several methods for balancing multiplayer games have been proposed in the past. One approach is to balance the configuration of game entities such as characters, weapons, or items~\cite{schreiber_game_2021}. To this end, the gaming industry uses pipelines to gain data-driven insights from games played and to adjust the balance on a regular basis\footnote{https://www.leagueoflegends.com/en-gb/news/dev/dev-balance-framework-update/}.

Evolutionary algorithms are further widely applied for game balancing.
Morosan et al. balance a real-time strategy game (RTS)~\cite{morosan_automated_2017} and Sorochan et al. generate new and balanced units for an RTS game~\cite{sorochan_generating_2022}. 
Volz et al. balance decks for a card game and, similar to this work, they evaluate the quality with a simulation-driven approach~\cite{volz_demonstrating_2016}.
De Mesentier Silva et al. optimize cards for the game Hearthstone~\cite{mesentier_silva_evolving_2019} to balance the metagame.
Pfau et al. train models with data from played games to replicate human behavior for automated entity balancing~\cite{pfau_dungeons_2020}.

%A different method, especially in online games, are matchmaking algorithms~\cite{alman_theoretical_2017}. These methods assign which players play against each other or are in the same team and contributes to balancing by ensuring that the players' skill in a match is at a comparable level. Multiple works on this exist, however, many are based on a form of the Elo rating system~\cite{elo_rating_1978} used by the World Chess Foundation where a player's skill can be expressed by a single number. To integrate more information into the process and also improve player satisfaction, match making in~\cite{deng_globally_2021} is framed as an MDP to apply RL in a sequential task.

A level's balance is furthermore affected by its design. This refers e.g., to the position of game entities in relation to the players' initial spawn position. As aforementioned, in many games this is ensured with (point) symmetric map architectures.
Contrary to the intended balance, players' perceptions may differ. A survey revealed that the prevailing opinion among highly experienced players is that only symmetric levels can actually be balanced~\cite{togelius_controllable_2013}.

To automatically construct balanced levels with PCG, several works using evolutionary algorithms have been introduced~\cite{lanzi_evolving_2014,lara-cabrera_balance_2014}. In contrast to this work, the fitness for balancing is evaluated using rule-based heuristics or metrics. These metrics are either hand-crafted or rely on a data-driven method using played games. Karavolos et al. extend this with a convolutional neural network to predict game related information which are then used in the fitness function of an evolutionary algorithm to produce balanced levels\cite{karavolos_using_2018}. A different approach is the usage of graph grammars~\cite{kowalski_strategic_2018}. The balance here is constructed by the rule-based placement of strategic game entities, which is, however, highly domain-dependent.
Rupp et al. present GEEvo, a framework for generating and balancing graph-based game economies~\cite{rupp_geevo_2024}. An evolutionary algorithm is used to balance the weights on the edges of the graph to a given value. Similar to this work, fitness is evaluated by running multiple simulations.
%In \cite{delaurentis_toward_2021} is a framework proposed to predict the balancing state based on played games with bots to balance a game.

% PCG
In this work, we focus on \emph{procedural content generation} for level generation, but the term PCG is also used in other applications such as the generation of game economies~\cite{rogers_using_2023,rupp_geevo_2024}, rules~\cite{cook_mechanic_2013} or narratives~\cite{alvarez_tropetwist_2022}. PCG methods span over from dedicated algorithms~\cite{mojang_mincraft_2011}, search-based methods~\cite{togelius_search-based_2011,lanzi_evolving_2014,lara-cabrera_balance_2014} up to the use of machine learning~\cite{summerville_procedural_2018} and deep learning~\cite{liu_deep_2021,giacomello_doom_2018,awiszus_toad-gan_2020,sudhakaran_prompt-guided_2023}. 
Once trained, machine learning methods can quickly generate content on demand, but they rely on the existence of game levels from which a model can be learned. As a result, these methods are usually not efficiently applicable to the creation of new games.

In the games community, RL has been widely applied to \emph{play} games~\cite{silver_general_2018, vinyals_grandmaster_2019}, but recently, it has also been applied to PCG~\cite{khalifa_pcgrl_2020,bontrager_learning_2021,gisslen_adversarial_2021,zakaria_procedural_2022}.
The framework we use and extend in this work is PCGRL~\cite{khalifa_pcgrl_2020}, utilizing RL for PCG tasks (cf. Section~\ref{sec:pcgrl}).
Zakaria et al. compare different deep learning-based PCG approaches for the puzzle game Sokoban, with uncontrollable PCGRL showing superior quality~\cite{zakaria_procedural_2022}.
There are already a few methods that have adapted the PCGRL method. In \cite{earle_learning_2021} controllable content generators were introduced, where users control the generated content with additional constraints, such as e.g., the number of players.
However, we do not compare our method to it, since our goal is only a balanced level.
Other approaches use evolutionary strategies on top of the PCGRL framework to achieve more content diversity~\cite{khalifa_mutation_2022} or adapt the method for 3D environments \cite{jiang_learning_2022-4}.

\section{Background}
\label{sec:background}
\subsection{The PCGRL Framework}
\label{sec:pcgrl}
In this work, we use the PCGRL \cite{khalifa_pcgrl_2020} framework for level balancing.
In PCGRL, PCG is formulated as a sequential decision-making task to maximize a given reward function, where semantic constraints can be expressed and thus, no training data is needed. To apply RL, the PCG problem is formulated as a Markov decision process (MDP).
Therefore, PCGRL introduces three different MDP representations for level generation. 
We will introduce three new swap representations based on the existing ones. Later, we will argue that they are more suitable for modifying game levels to improve the balance state. The original representations in PCGRL are: 

\paragraph{Narrow} This representation randomly selects a tile in the grid and the agent only has to decide what type of tile to place on the selected position. The action space is small as it only consists of the different tile types.

\paragraph{Turtle} The \emph{turtle} representation allows the agent to move around the map and then to decide what type of tile to place on the current position. The advantage of the narrow representation is that the agent is not restricted to the randomly assigned position and can therefore learn where to move next.

\paragraph{Wide} The \emph{wide} representation gives the agent full control over the level generation process, since the agent can decide which tile of the entire grid should be changed. This greatly increases the action space as each position of the grid multiplied by the number of tiles represents an action. Because the agent can change everything directly according to a plan it has constructed, it is the most human-like representation.

\subsection{The Neural MMO Environment}
\label{sec:nmmo}
As a basis for our experiments, we use the NMMO environment~\cite{suarez_neural_2019}. The NMMO environment is a Massively Multiplayer Online Battle Royal-inspired multi-agent RL environment for research.
It simulates a tile-based virtual world in which agents must forage for resources to survive and prepare for battles with other agents. All agent actions are processed simultaneously per time step. The agent that survives the longest wins the game.
The state of an agent is represented by its position as well as its current health, food, and water levels.

In this paper, we focus on the balancing in the PCG process of the game map. To apply and evaluate our method, we simplify the winning condition and frame the game as a forage survival game. Therefore, we disable the combat system and restrict the map tiles to the types of grass~\raisebox{-0.2em}{\includegraphics[width=1em]{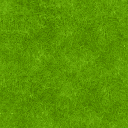}}, forest~\raisebox{-0.2em}{\includegraphics[width=1em]{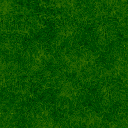}}, stone~\raisebox{-0.2em}{\includegraphics[width=1em]{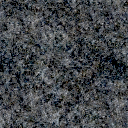}}, and water~\raisebox{-0.2em}{\includegraphics[width=1em]{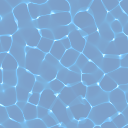}}. Agents~\raisebox{-0.2em}{\includegraphics[width=1em]{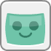}} cannot move on stone or water tiles. Forest and water are resources that the agent can gather. 
%A detailed overview of the tiles and their behaviors is given in Figure~\ref{fig:tile-descr}.
When moving on a forest tile, it is automatically consumed as food and refills the agent's food indicator. In the process, the forest tile changes to the state \emph{scrub}. Scrub tiles cannot be consumed, but there is a 2.5\% chance per time step per scrub tile that it will respawn and transit back to the forest state.
To refill an agent's water indicator, it must simply step on an adjacent tile. Water tiles are not depleted. If one of both resource indicators are empty, the agent loses a fixed amount of health per step. If the agent's health indicator reaches zero, the agent dies. An agent can regain health if its water and food indicators are above 50\%. 

%\begin{figure}
%     \centering
%     \includegraphics[width=0.7\linewidth]{content/images/tiles.png}
%     \caption{Description of the NMMO tiles.}
%     \label{fig:tile-descr}
%\end{figure}

For this research, the winning conditions are the same for both agents. The first agent to reach one of the two goals wins:
\begin{itemize}
    \item Collect five food resources.
    \item Last agent standing: Neither starved nor died of thirst.
\end{itemize}

In this research, we fix the map size to a grid of 6x6 tiles and two players to speed up the experiments and to keep the results interpretable by human evaluators. Furthermore, the smaller grid size limits the available space and resources, which increases the competitive race of the agents. This competitive race is to be balanced in this paper.

\section{Method}
\label{sec:method}
\subsection{Balancing Architecture}
\label{sec:architecture}

\begin{figure*}
  \centering
  \includegraphics[width=0.9\linewidth]{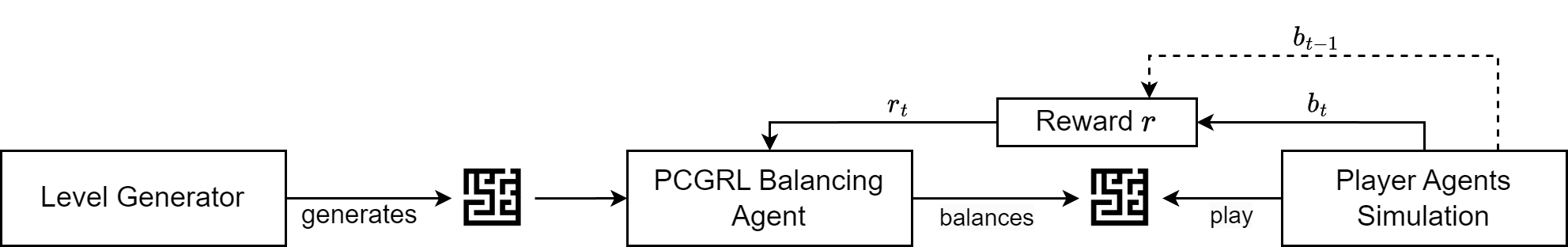}
  \caption{Description of the balancing architecture. It is separated into three units: A level generator, a level balancing agent, and a game playing simulation. In the latter, the game is simulated by playing it $n$-times with heuristic player agents. The reward $r_t$ for training the balancing agent is computed from the balance states $b_t$ and $b_{t-1}$ of the simulations.}
  \label{fig:architecture}
\end{figure*}

This paper proposes an architecture for balancing tile-based levels with RL. The idea is to reward a PCGRL agent for only altering a given playable level towards a balanced level.
A balanced level is one where all player agents with the same skill have the same win rate.
We focus only on game balance through level design, which excludes balancing a sequence of levels or different play styles for instance.
The architecture is shown in Figure~\ref{fig:architecture}. It is separated into three parts: a level generator, a balancing agent, and a game playing simulation.

% short descr of gen agent
The level generator constructs a playable level from random noise which is then fed to the balancing agent. In this work we use PCGRL for the generator, but other PCG methods will serve comparable at this point. The generator is trained separately before training the balancing agent.

% short descr of balancing agent
The core idea of level balancing is not to generate a new level, but to modify the given level to satisfy the balancing constraint.
At each time step, the balancing agent can decide to swap the positions of two selected tiles. If a swap (cf. Section \ref{sec:swap}) has been made, the level is played $n$-times in a simulation by player agents.
Subsequently, the balancing agent is rewarded based on how this action affected the balance state in simulations (Section \ref{sec:reward}). 
The simulations can therefore be understood as the basis for a static simulation-based evaluation function as described by Yannakakis and Togelius~\cite{yannakakis_experience_2011}.
More details are given in the implementation details~(Section~\ref{sec:details}).

\subsection{Swap-based Representation Pattern}
\label{sec:swap}
To formulate the problem of PCG with RL as MDP, three representations have been introduced in \cite{khalifa_pcgrl_2020}. In all of them, the agent can decide to replace a tile at a particular position. This method, however, can lead to unplayable levels at a time step. Furthermore, to move the position of e.g., a player tile somewhere else, the agent would first have to remove the player tile before creating it at a different position. In this time step, the level would not be playable for the player agents, and thus, no reward could be computed in the simulation step. Additionally, the agent would first receive a negative reward due to the number of players is now invalid. The subsequent creation at a different position under the previously given negative reward is hard to learn for RL agents.

For this reason, we introduce a swap-based representation pattern. In these representations, the agent can decide to swap the positions of two tiles per time step.
Not adding or removing tiles entirely is a more robust approach to ensuring level playability.

However, there may be game domains where multiple tiles have a semantic connection, such as multiple river tiles forming a river. Swapping these tiles around can break playability. Therefore, in the balancing step, we suggest sending unplayable levels back to the generator for repair.
This repair is similar to the level fixing seen in~\cite{siper_path_2022}, but it would also involve an additional increase in computation.

In this work, we demonstrate the power of swapping with a simpler domain where there are no semantic relationships between multiple tiles. In addition, swapping tiles ensures that the old level isn't simply regenerated from scratch, since that is the job of the level generator.
Swapping two tiles of the same type has no effect on balance. Thus, we prevent these swaps to reduce the computational effort. In these cases, the agent is rewarded with 0.
Therefore, we extend the narrow, turtle and wide representations~\cite{khalifa_pcgrl_2020} with a swapping mechanism. Like in PCGRL, observations are one-hot encoded in all representations. The detailed description for each swap representation is provided below.

\paragraph{Swap-Narrow} At each time step two random tile positions are presented to the agent, and it can decide to swap the tiles or not. The agent's limited positional control results in a very small action space $A$ with only two actions: swap or do not swap. $A$ is therefore $A=[2]$.
\paragraph{Swap-Turtle} Starting from two random positions, the agent can decide to swap the tiles at the current positions in each time step. If no change is made, it can decide to which adjacent tile to move to next. As in the original PCGRL, staying at a position and not changing a tile is not an option. $A$ is therefore $A=[4,4,2]$, which results in 32 possible actions.
\paragraph{Swap-Wide} In this representation, the agent sees the entire level and can freely determine the tile positions and whether to swap them. It can be interpreted as looking at the whole level and then deciding what to move where. A drawback here is the large action space, since it scales twice with the grid width $w$ and height $h$. $A$ is therefore $A=[w,h,w,h,2]$. In the case of a square grid, as in this work with a size of 6, $A$ has 2592 possible actions.

\subsection{Reward Design}
\label{sec:reward}
The reward function is crucial for the successful training of an RL agent. Heuristic approaches for evaluating the balance have been introduced in~\cite{lanzi_evolving_2014,lara-cabrera_balance_2014}. These approaches, however, contain domain-specific information and are thus not transferable. To address this shortcoming, we propose the use of a more generic, game domain-independent reward function that depends only on how often each player wins. In addition, we require that the game has at least one winner at the end. Since draws also provide information about the balance, our reward function must take this into account as well.

To adequately measure the balance, we are inspired by fairness metrics from the fair machine learning community. A survey of common metrics used in that field is provided by Makhlouf et al.~\cite{makhlouf_applicability_2021}. The statistical parity metric~\cite{dwork_fairness_2012} is one of them and is often used to measure fairness between two groups. The fairness for two groups $G=1$ and $G=2$ is considered as fair if the conditional probability of the same outcome $\hat{Y}$ is the same for both groups, as seen in Eq.~\ref{eq:stat-parity}. 

\begin{equation}
\label{eq:stat-parity}
P(\hat{Y}\,|\,G = 1) = P(\hat{Y}\,|\,G = 2)
\end{equation}

In this work, we transfer this definition of fairness to the domain of game balancing. 
In the context of the statistical parity, the two groups represent the two players, and we want to optimize their win rates.
%, cf. $P(win \,|\, player = 1) = P(win \,|\, player = 2)$. 
However, the direct use of Eq.~\ref{eq:stat-parity} is not applicable to RL since it requires a function that rewards the agent with positive or negative incentives per time step $t$. In addition, an intermediate reward design which compares $t$ to $t-1$ has shown good results in PCGRL.
First, we therefore define the win rate $w_{p_it}$ per player $p_i$:
\begin{equation}
\label{eq:bt}
w_{p_it} = P(w_t\,|\,p = i)
\end{equation}
$w_{p_it}$ is then inferred from the data of the $n$-runs of the player agent simulations per step. This can be seen as a sampling from the actual win rate distribution (cf. Section~\ref{sec:simulation} and Fig.~\ref{img:balancing}).
Second, in this extended paper, we additionally want to make the reward design configurable to a certain balance state $b$ and thus rewrite the balancing calculation $b_t$ as:

\begin{equation}
\label{eq:b}
b_t = |\, w_{p_1t} - b \,|
\end{equation}

%The balancing state $b_t$ is the absolute value of the deviation of a player from $b$.
Due to the use of the absolute value, $w_{p_it}$ of both players could be used in Eq.~\ref{eq:b}, since their values describe probabilities that sum up to 1.
%We design the reward function to reward the agent for improving the balancing state $b_t$ in time step $t$ in comparison to $b_{t-1}$.
$b_t$ is defined in $[0,1]$ as well, where 0 indicates that $b_t$ is exactly equal to $b$; 1 indicates the maximum deviation from $b$. In the example where an exact balance between both players is desired ($b=0.5$), $b_t=1$ indicates a particular player wins every game.
With $b_t$ in this range, there is flexibility to configure towards which player the game should be balanced against.
Finally, the reward $r_t$ is: 
\begin{equation}
\label{eq:reward}
r_t = b_{t-1} - b_t  + \alpha
\end{equation}

To reward the agent additionally, a reward $\alpha$ is given if $b_t$ is exactly $b$. Otherwise, $\alpha$ is 0. If $\alpha>0$, the episode always ends since $b_t=b$.
As a result, the reward will be positive if the agent improves the balance state, negative otherwise. For no impact on the balance state, the agent receives no reward (value 0). Using this reward design, the RL agent is gradually incentivized to reduce the absolute difference from the current balance state to the desired one.

\section{Implementation Details}
\label{sec:details}

\subsection{Level Generator}
\label{sec:gen}

The task of this unit is to generate playable levels for balancing. In this work, the generator is a model trained using the PCGRL framework.
The reward for the training process is designed to reward the agent for having exactly two players and creating a valid path between them.
To ensure direct competition, the latter restriction ensures that both players have access to the same area of the game level. Additionally, it prevents single players from being locked behind stone walls.

The NMMO environment uses PCG by default to generate random levels, however, the players' spawn positions are randomly determined at the start of the game. Nevertheless, the spawn positions are crucial for the balance, so the balancing agent should be able to influence them. For this reason, player tiles representing these positions are integrated in the level generation process.

If there is a valid path between both players, the agent receives a positive reward, otherwise a negative one. Additionally, at each step the agent is rewarded with the difference of the players that are and should be (similar to \cite{khalifa_pcgrl_2020}). We get the best results using the \emph{wide} representation. An episode ends when both constraints are met, or when the agent exceeds a fixed number of permitted steps or changes.

After training, the model can produce levels satisfying the given constraints in 98,7\% levels out of an evaluation sample of 5000. The generated levels achieve a maximum diversity at 100\%, so no two levels are completely identical. An example of a generated level is shown in Figure~\ref{fig:gen-level1}~(left). The functionality of the game and tiles is given in Section~\ref{sec:nmmo}.

Level generation in this step continues until the agent has produced a valid level that satisfies the constraints to ensure that the balancing agent receives playable levels only. The generated levels are then passed on to the balancing agent.
%\begin{figure}[!t]
%  \centering
%  \label{img:genagent}
%  \includegraphics[width=2.5in]{content/images/pcgrl_map.png}
%  \caption{A playable, not yet balanced game level for the NMMO environment generated with PCGRL. Exactly two players exist which are connected. The colored tiles represent the NMMO tiles: blue=water, grey=stone, light green=grass and, dark green=forest. Additionally, the spawn positions of the players are represented with the face icon.}
%\end{figure}

\subsection{Balancing Agent}
\label{sec:balancing}
The balancing agent is the core component of the architecture in Figure \ref{fig:architecture}. We model the agent as a PCGRL agent that can decide to alter a previously generated game level per time step. To support this process, we use swapping representations.
The reward is computed from the results of a simulation of the balance state (Section \ref{sec:reward}).

The observation is the current level one-hot encoded. In detail, this depends on the chosen representation. As RL algorithm, we use PPO (proximal policy optimization)~\cite{schulman_proximal_2017}.
An episode ends when the level is either balanced, or a fixed number of steps or changes is exceeded. In PCGRL, the number of changes allowed is determined relative to the grid size (cf. change percentage). As in the original PCGRL we use a change percentage of 20\%. We set the value for the number of allowed steps within an episode to 100.

\subsection{Player Simulation}
\label{sec:simulation}
To compute $b_t$ for the reward (Section~\ref{sec:balancing}), we run a simulation $n$-times of player agents playing the game.
The player agents can be any solution which can simulate a player's behavior with a desired quality. In this work we use the scripted \emph{Forage}-agent for the NMMO environment, which is publicly available\footnote{https://github.com/NeuralMMO/baselines}. An agent gathers the nearest available food resource. If its water indicator drops below a certain threshold, it will search for the nearest water resource to refill it.
The usage of scripted agents provides the advantage of a deterministic behavior at each time step. 
At this point, however, the use of trained agents with e.g., RL, different skilled or types of agents would also be feasible. See the discussion~(Section~\ref{sec:discussion}) for further information.

Despite the simulation is run with the same player agents, the winners may differ each pass. This is due to the probabilistic mechanics that make games interesting, such as dice rolling or the respawning of resources with a certain probability. A player can win with luck, but when playing many times, skill should make the difference. The question then becomes, how often must a simulation be run to minimize the noise of the probability? Since the simulations are computationally intensive, it is of interest to find the number of minimum runs where the win rates vary at an acceptable level.

\begin{figure}[!t]
  \centering
  \includegraphics[width=3.1in]{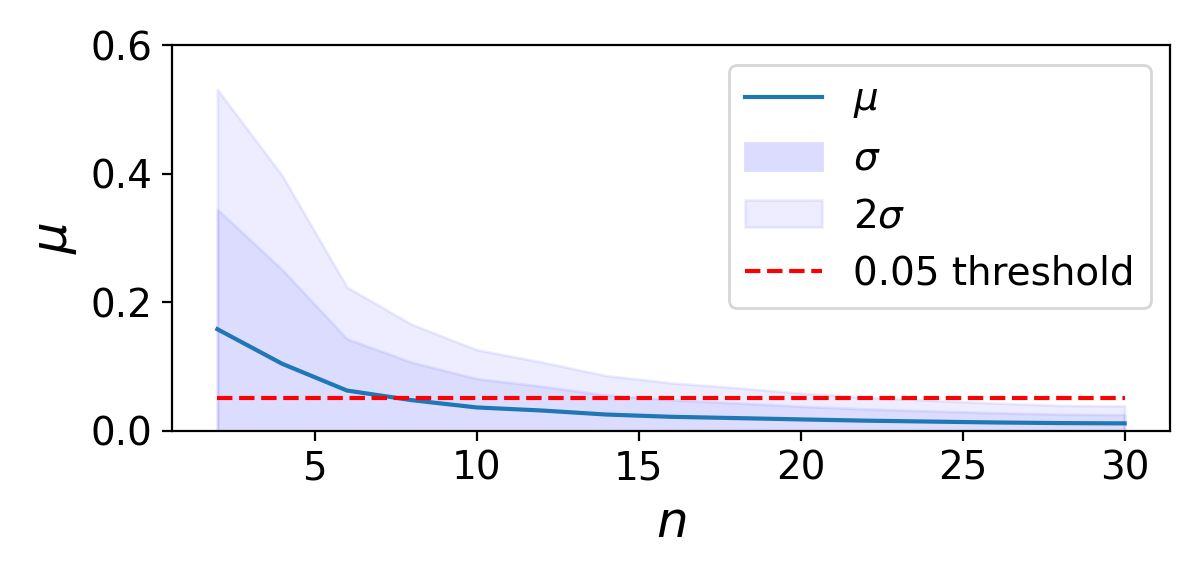}
  \caption{How many times $n$ should we run the simulation to approximate the balance state? We figure that out by calculating the mean deviation $\mu_n$ of win rates from $n-2$ to $n$ for the investigation of fluctuations in win rates.}
  \label{img:balancing}
\end{figure}

We approach this by investigating how much the win rates $w_n$ of a given number of simulations $n$ differ on average from $n-2$. Only even values for $n$ are applicable since otherwise a balanced game is not possible. Therefore, we run the simulation on a sample of size $s=500$ levels up to a number of $N=30$ times and compare how the win rates for $n$ vary on average in comparison to $n-2$. The average deviation~$\mu_n$ of a particular $n$ is expressed with: $\mu_n = \frac{1}{s} \sum^{N}_{n=0} | w_n - w_{n-2} |$ and its course is shown in Figure~\ref{img:balancing}. In addition, the standard deviation is given with $\sigma$ and $2\sigma$. It is clear that the larger $n$, the smaller $\mu_n$. To determine a reasonable $n$, we set the threshold: $\mu_n + \sigma < 0.05$. For this game with the configured number of players, this is $n \ge 14$. Therefore, we use $n=14$ in this work.
This method can be used to determine $n$ for any domain.

\section{Experiments and Results}
\label{sec:experiments}

We evaluate our architecture in several steps: First, we sample a fixed dataset of levels to use for direct comparison with the generator. Second, we compare the balancing performances of the three introduced representations with the original PCGRL method as a baseline~(\ref{sec:eval-swap}). To generate balanced levels with the original PCGRL method, we integrated the balancing constraint into the reward function.

Subsequently, we investigate on the levels the models created~(\ref{sec:gen-levels}) and examine which tiles the models swapped in the level altering process~(\ref{sec:impact}). The latter can provide insights into which tiles actually affect the balance.

\subsection{Performance Overall}
\label{sec:eval-swap}
We compare the performance of the swap representations against each other and against the original PCGRL as baseline\footnote{Only the narrow representation is shown here; the other two give similar results.}. For direct comparison, all models were trained with the same number of 200k training steps, resulting in 970 policy updates.

\begin{figure}[!t]
  \centering
  \includegraphics[width=2.9in]{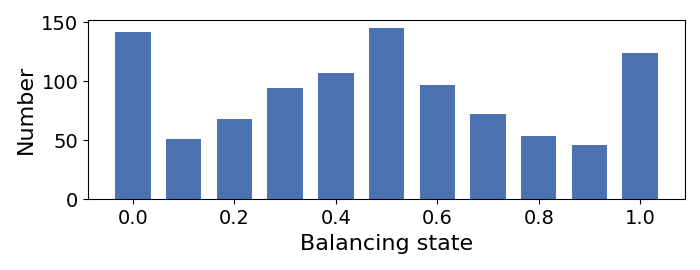}
  \caption{Distribution of the initial balance states based on the players' win rates in the generated dataset of 1000 levels. We use this dataset to compare the different representations.}
  \label{fig:init-balancing}
\end{figure}

For the evaluation, we create a dataset of 1000 levels with our PCGRL generator. This dataset is then used for all four models for evaluation.
First, we examine the distribution of the initial balance states of the levels in the dataset (see Figure~\ref{fig:init-balancing}). It is important to note that the distribution is not uniform. Except for the states 0 and 1, the levels seem to be normally distributed around the state of 0.5, indicating that both players won equally often. Initially, 13.6\% of the levels are balanced. However, peaks towards maximally unbalanced levels can be observed at the outer edges of the distribution. Maximally unbalanced levels make up 26.6\% of the dataset.

To evaluate the performance of the balancing method we compare the balance state before and after per representation. A general overview is given in Table~\ref{tab-overview}. A histogram of the balancing improvement is shown in Figure~\ref{fig:repr-compare}. 

\begin{figure}
     \centering
     \begin{subfigure}[b]{0.5\textwidth}
          \centering
          \includegraphics[width=\textwidth]{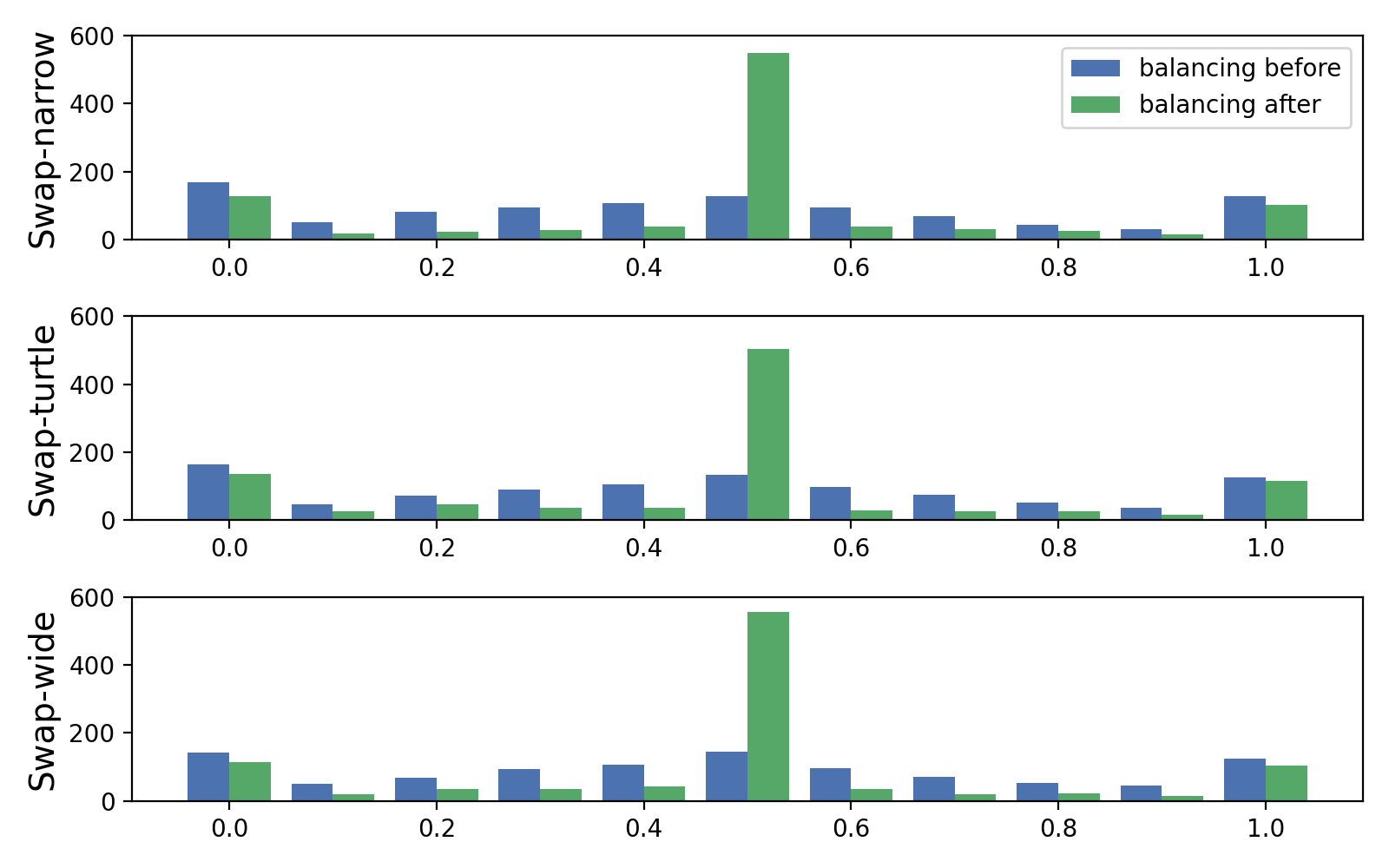}
          \caption{\normalfont Swap representations.}
          \label{fig:repr-compare}
     \end{subfigure}
     \vspace{\fill}
     \begin{subfigure}[b]{0.5\textwidth}
         \centering
         \includegraphics[width=\textwidth]{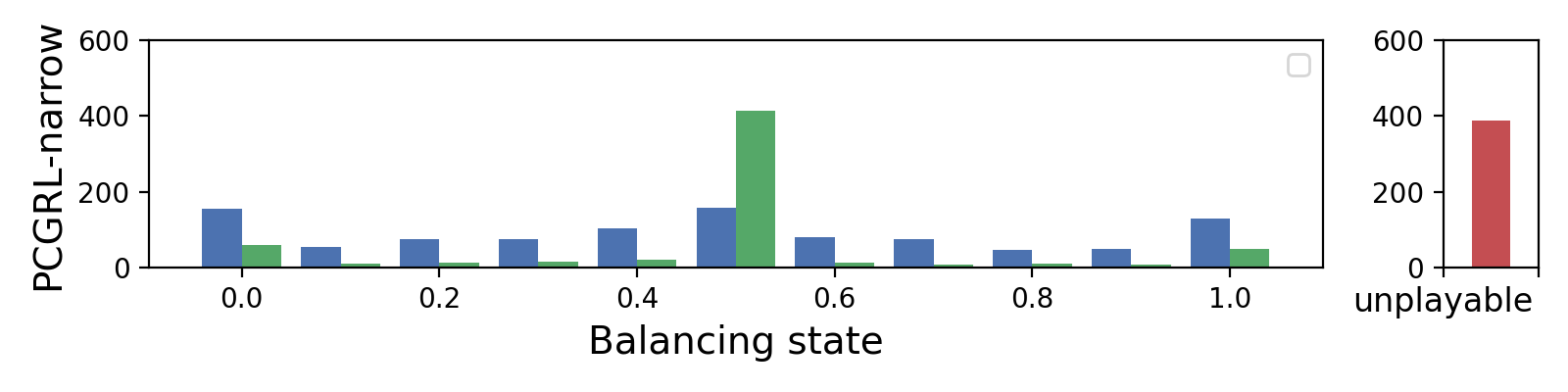}
         \caption{\normalfont Original PCGRL (narrow).}
         \label{fig:narrow-pcgrl}
     \end{subfigure}
          \caption{Comparison of the balance state distributions before and after the balancing process for each representation using the dataset of 1000 levels (Figure~\ref{fig:init-balancing}). The swap representations (a) are compared to the original PCGRL implementation directly~(b).}
     \label{fig:repr-compare}
\end{figure}

The performance of all three swap representations is of comparable quality. Each representation managed to improve the proportion of balanced levels. The swap-narrow and swap-wide representations perform slightly better than swap-turtle in terms of balancing.
The original PCGRL narrow representation also improved the share of balanced levels, however, with 36.7\% this share is smaller compared to the swap representations. Additionally, a proportion of 38.8\% of the levels are in an unplayable state at the end of an episode~(Figure~\ref{fig:narrow-pcgrl}); whereas our methods result in 100\% of playable levels in this domain in all cases.
However, unbalanced levels remain in all results. The largest proportion remains for the balance states 0 and 1.
For representations with a higher proportion of balanced levels, the average episode lengths are shorter. This is because the remaining unbalanced levels maximize the episode length.

%\begin{figure}[!t]
 % \centering
 % \includegraphics[width=3.5in]{content/images/repr_compare.png}
 % \caption{Comparison of balancing distributions of the balancing states before and after the balancing method per each representation. The same dataset of of 1000 levels is used each time.}
%  \label{fig:repr-compare}
%\end{figure}

%\begin{table}[]
%\centering
%caption{Performance overview of the representations. Initially balanced levels were not taken into account.}
%\label{tab-overview}
%\begin{tabular}{@{}lllll@{}}
%\toprule
%                        & Swap-narrow  & Swap-turtle   & Swap-wide    & PCGRL        \\ \midrule
%Balanced (\%)           & \textbf{48.1}   & 42.5          & \textbf{48.1}  & 30.4    \\
%Improved (\%)           & 63.3         & 56.8          & \textbf{64.1}    & 36.7      \\
%Playable (\%)           & 100          & 100           & 100            & 61.2      \\
%Avg. changes            & 4.6$\pm$1.9  & 4.7$\pm$1.9   & 4.9$\pm$1.9  & 5$\pm$1.8    \\
%Avg. ep. length         & 11.3$\pm$6.1 & 25.4$\pm$17.7 & 14.1$\pm$7.9 & 15$\pm$7.9   \\
%Size action space       & 2            & 32            & 2592         & 10           \\
%Compute time (h)       & \textbf{52}   & 40              & 33            &  \\
%\bottomrule
%\end{tabular}
%\end{table}

\begin{table}[btp]
\caption{Performance overview of the representations. Initially balanced levels were not considered.}
\begin{center}
\begin{tabular}{lllll} \toprule
                        & {\textbf{S-narrow$^{\mathrm{*}}$}}  & {\textbf{S-turtle$^{\mathrm{*}}$}} & {\textbf{S-wide$^{\mathrm{*}}$}} & {\textbf{PCGRL}}   \\
\cmidrule(lr){1-1} \cmidrule(lr){2-2} \cmidrule(lr){3-3} \cmidrule(lr){4-4} \cmidrule(lr){5-5}
Balanced (\%)           & \textbf{48.1}   & 42.5          & \textbf{48.1}  & 30.4    \\
Improved (\%)           & 63.3         & 56.8          & \textbf{64.1}    & 36.7      \\
Avg. changes            & 4.6$\pm$1.9  & 4.7$\pm$1.9   & 4.9$\pm$1.9  & 5$\pm$1.8    \\
Avg. ep. length         & 11.3$\pm$6.1 & 25.4$\pm$17.7 & 14.1$\pm$7.9 & 15$\pm$7.9   \\
Size action space            & 2            & 32            & 2592         & 10           \\ \bottomrule
%\multicolumn{2}{l}{} \\ 
\vspace{-2mm} \\
\multicolumn{4}{l}{$^{\mathrm{*}}$Swap representations are abbreviated with S.}
\end{tabular}
\label{tab-overview}
\end{center}
\vspace{-3mm}
\end{table}

\subsection{Generated Levels}
\label{sec:gen-levels}
\begin{figure}
     \centering
     \begin{subfigure}[b]{0.41\textwidth}
         \centering
         \includegraphics[width=58mm]{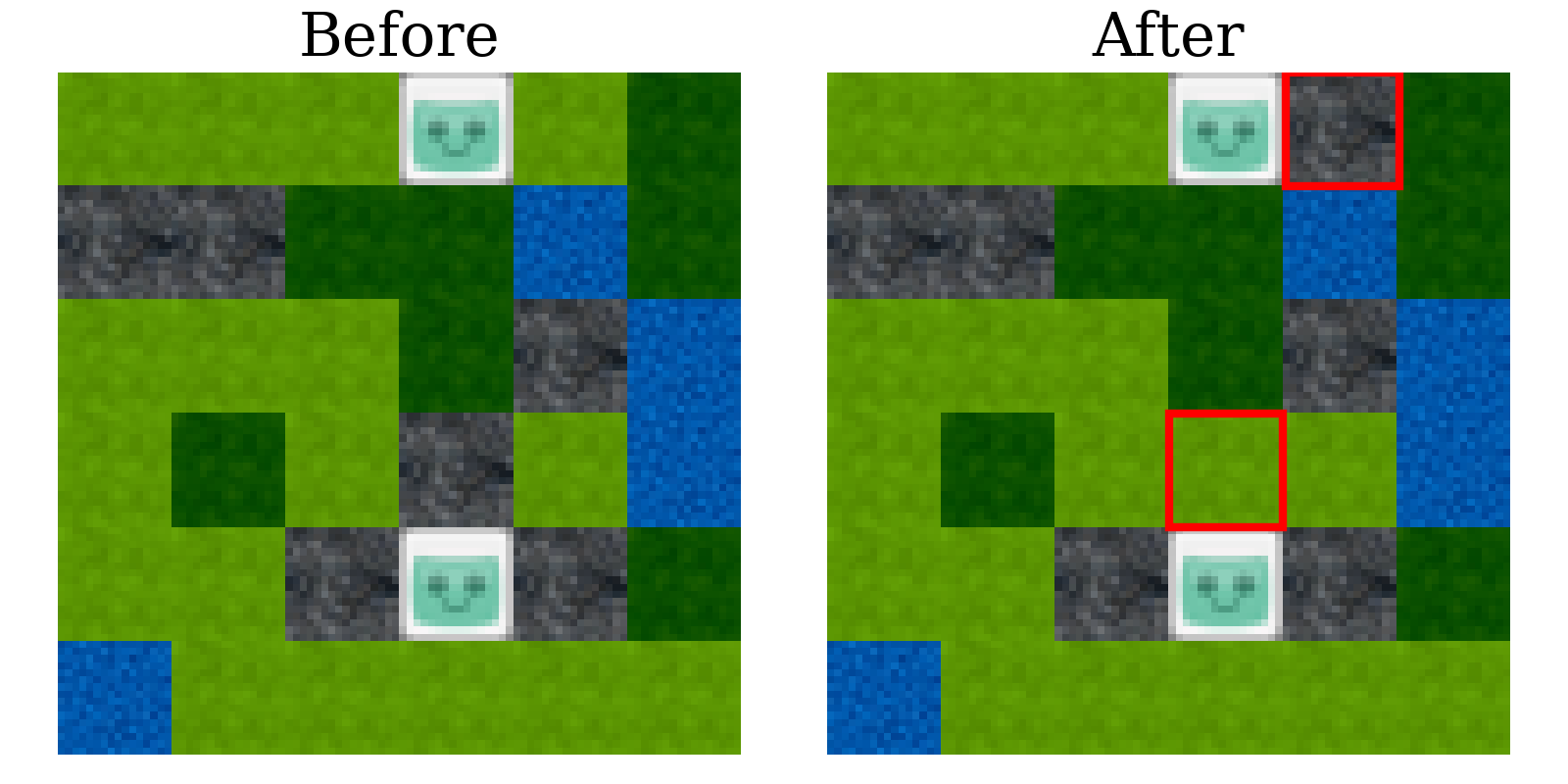}
         \caption{\normalfont The agent balanced a maximum unbalanced level (0) to a balanced level (0.5) by swapping one tile.}
         \label{fig:res-perfect}
     \end{subfigure}
      \vspace{\fill}
     \begin{subfigure}[b]{0.41\textwidth}
         \centering
         \includegraphics[width=58mm]{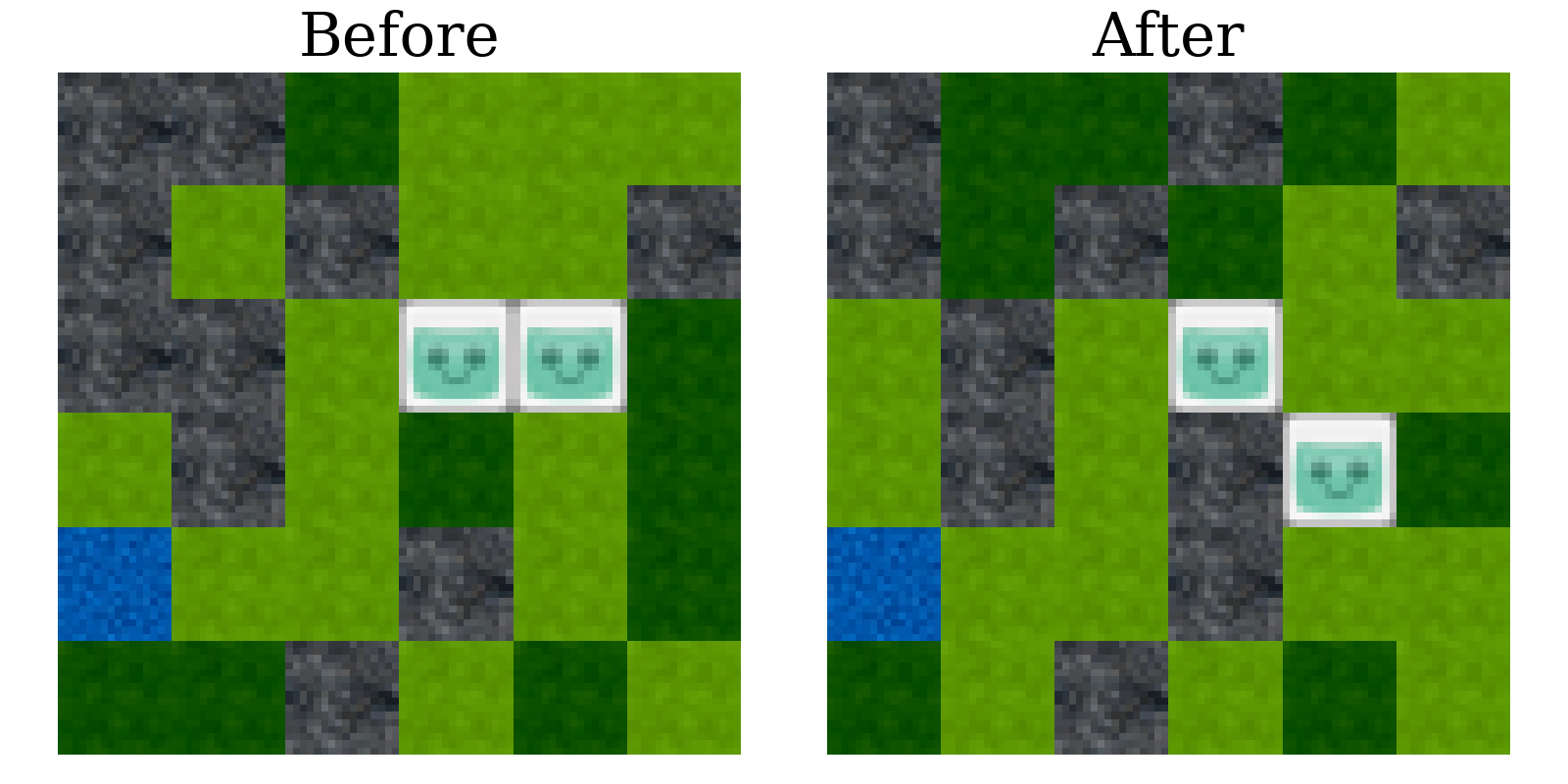}
         \caption{\normalfont The agent could not change the initial balance of 0. The generation stopped after exceeding the allowed number of changes. Since several tiles were swapped, they are not highlighted for the sake of visualization.}
         \label{fig:res-failed}
     \end{subfigure}
    
    \caption{Examples of levels modified by the balancing agent. The left column shows the generated levels before balancing. The right column levels are the results after the balancing. Swapped tiles are marked with red frames.}
     \label{fig:gen-levels}
\end{figure}

Figures \ref{fig:gen-level1} and \ref{fig:gen-levels} show examples of the different types of generated levels taken from the samples in Section \ref{sec:eval-swap}. Figure \ref{fig:res-perfect} is an example where the balancing agent altered the given level to a balanced level by swapping only one tile. By swapping the highlighted grass tile with the rock tile the path to the resource (forest) tiles is now blocked. This results in a more equal availability of food resources for both players. In Figure \ref{fig:gen-level1}, the agent has balanced the level from an initial balance state of 0.3 to 0.5. By swapping the marked water tile to a more central position for both players, the balance is improved. In Figure \ref{fig:res-failed} the agent failed to balance the initial level with $b_0=0$. The balancing process terminated after reaching the maximum number of changes permitted. In the end, $b$ is still 0.

\subsection{Impact of Tiles on Balancing}
\label{sec:impact}
The analysis of the actual swaps made by a model gives insight into its behavior. Thus, we have shown that the model could improve the balance state of the given levels, we can further argue that the swaps made by the model have an impact on the balancing. Conversely, we can infer from this behavior which tiles in the game have the most impact on the balance.

\begin{figure}[!t]
  \centering
  \includegraphics[width=3.5in]{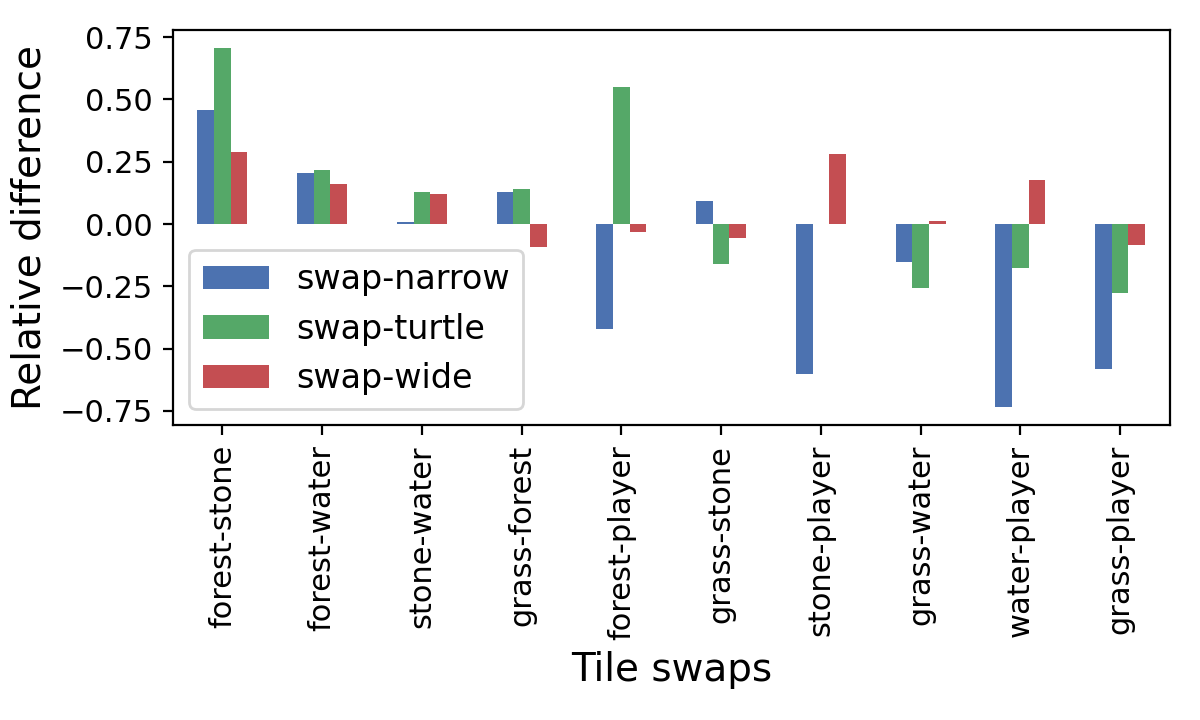}
  \caption{Comparison of the swapped tiles by the model per representation on the generated 1000 levels. The comparison is made with respect to the inverse tile type distribution of all levels. This shows the difference to random swapping.}
  \label{fig:swaps}
\end{figure}

For comparison, for each pair of swaps, we compute the relative differences between the frequencies of random swaps and those of the models. These frequencies are then factorized with the inverse probabilities of the total tile occurrences in the dataset. Due to swapping the same tile types is prohibited by the representations, ten different combinations are possible. This is shown in Figure~\ref{fig:swaps}.
The Figure shows that the three representations have different behaviors, but they agree on particular points. Swapping forest for stone tiles has the most impact, followed by swapping forest for water tiles. This makes sense in the context of the resource gathering win condition. Moving the resource tiles forest or water is likely to affect the balance as they are included in the win condition.
A stone tile can be used to block a player's path. Therefore, swapping forest tiles for stone tiles has an additional powerful effect.
Surprisingly, swapping a player's spawn position with other tiles is not a favored action in all cases.

\section{Extended Study on Overall Improvements and Balancing with a Varied Target Balance}
\label{sec:extensions}

In Section~\ref{sec:experiments}, we focused on a balance where both players win equally often. This study extends the core idea of the paper by showing that this method can also optimize given levels towards arbitrary balancing values. Furthermore, the performance of the trained models is improved.

\subsection{Improvement of Results}
\label{sec:improvements}

To improve the results of Section~\ref{sec:experiments}, we addressed two issues:
First, both players' spawn positions were previously represented by the same tile type, making it impossible for the model to distinguish between them.
For this reason, we now assign different tile representations to each spawn position, as our previous research has shown that these affect balance.
To differentiate when drawing, we assign green to player 1~\raisebox{-0.2em}{\includegraphics[width=1em]{content/images/tiles/player.png}} and yellow to player 2~\raisebox{-0.2em}{\includegraphics[width=1em]{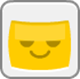}}.
The different encoding of the two positions adds more information and thus slightly increases the observation space, but the action space remains the same. Training under identical conditions as described in Section~\ref{sec:experiments} improves the previous results to 65.6\% of levels reaching balance for $b=0.5$.
Second, we improved the performance of our code, reducing computation time. As a result, training became more efficient, allowing for more updates (10k) to the model's policy in the same amount of time.

In the results of the experiments in Table~\ref{tab-overview}, the best model could balance 48.1\% and could improve 64.1\% of the levels to a balance of 0.5. In this extended work, we improve the proportion of balanced levels to 68.0\% compared to the baseline (Table~\ref{tab:unbalancing}). The extended training could additionally lead to a slight improvement in performance (2.4 percentage points).
Levels where the model could at least improve towards the desired balancing are improved to a proportion of 88.9\%. A level is considered improved if the balance is equal to or comes closer to the desired state.

\subsection{Balancing with Different Target Values}
\label{sec:unbalancing}
With this set of experiments, we extend the previous research by also training models to generate levels that are not equally balanced for both players. This shows that our method is generally capable of balancing levels to a certain value.

Therefore, we use the adapted reward function (Eq.~\ref{eq:reward}), which is configurable with a parameter $b$ towards a desired balance. We investigate this by training six different models in tenth increments of possible values for $b$ in the interval $[0,0.5]$. As it performed best according to the previous results we use the swap-wide representation in combination with the improvements from Section~\ref{sec:improvements}. We refrain from implementing a single but controllable approach as in~\cite{earle_learning_2021}, as we are only experimenting with different parametrizations of the reward function. Applying controllable RL would imply an adaptation of the previously defined MDP, making a direct comparison with the previous results difficult.
To evaluate all models, we use the same level dataset from Section~\ref{sec:eval-swap} to ensure comparability. An overview of the results is given in Table~\ref{tab:unbalancing}.
%Figure~\ref{fig:unbalancing}.

For all values of $b$, a significant improvement of the respective balance state can be observed (blue). The performance, however, differs depending on $b$. For $b \geq 0.3$, the proportion of levels with the desired balance is even higher than the baseline from the previous experiments. For the value 0, which indicates levels that are heavily balanced towards a particular player, the performance is slightly worse than the baseline. In all cases, however, the models were able to improve the balancing towards the given value.

%The creation of diverse content is of interest when implementing a PCG method~\cite{togelius_search-based_2011}. For all generated levels per model, we therefore evaluate their diversity by matrix comparisons. For each $b$, we determined a content diversity of 100\% of the balanced levels. For more information on level asymmetries, see Section~\ref{sec:asymmetries}.
%This \hl{work} aims to generate asymmetrical yet balanced levels. To evaluate this, we checked all levels for each model, considering both diagonals as well as the vertical and horizontal aies. None of the levels created had any form of symmetry.
%Both, the checking of diversity and symmetry are solutions at the value layer, however, not at the semantic one. There may be levels that are semantically the same e.g., if only a few tiles are different, which have no influence on the game. This process, however, cannot be easily automated.

\begin{table}
    \centering
    \caption{Performance overview of six swap-wide models trained to optimize different balance values $b$ over the interval $[0,0.5]$ of the balance metric (cf. Section~\ref{sec:reward}). All values are given as a percentage.}
    \begin{tabular}{ccccc} \toprule
         $b$ &  \textbf{Balanced to $b$}&  \textbf{Improved towards $b$} & \textbf{Initial $=b$} & \textbf{Baseline} \\
\cmidrule(lr){1-1} \cmidrule(lr){2-2} \cmidrule(lr){3-3} \cmidrule(lr){4-4} \cmidrule(lr){5-5}
         
         0.0&  36.1&  71.1&  8.6& -\\
         0.1&  35.1&  73.5&  7.2& -\\
         0.2&  42.9&  76.3&  6.4& -\\
         0.3&  54.6&  77.9&  11.4& -\\
         0.4&  60.7&  81.4&  12.0& -\\
         0.5&  \textbf{68.0} &  88.9&  15.6& 48.1\\
         \bottomrule
    \end{tabular}
    \label{tab:unbalancing}
\end{table}

%\begin{figure*}[!t]
%  \centering
%  \includegraphics[width=0.75\linewidth]{content/images/unbalancing_overview.png}
%  \caption{Performance overview of eleven swap-wide models trained to optimize different balancing values $b$ across the interval $[0,1]$ of the balancing metric (cf. Section~\ref{sec:reward}). The plot shows the improvement of the proportion of the levels which are initially in the desirable state (red) to the proportion after the balancing proceeding (blue). The green line represents the proportion of levels where the models could at least improve the balancing value towards the desired goal. For all results, levels which are already in the desired balancing state have been filtered out. As a baseline, the previous result from Section~\ref{sec:experiments} is shown~(orange).}
%  \label{fig:unbalancing}
%\end{figure*}

%\section{Extended study on adding more player agents}

%\subsection{Reward design for n-players}
%\begin{equation}
%\label{eq:new-bt}
%wr_{p,t} = \frac{1}{wl_t} \sum_{i=1}^{w_{p,t}} w_{p,t,i}
%\end{equation}

%\begin{equation}
%\label{eq:new-bt}
%b_t = \sum_{p=1}^{P} | wr_{p,t}-\frac{1}{P} |
%\end{equation}

%\subsection{Level generator for N players}

%\subsection{Direct comparison of the old reward function with new function}

%\subsection{Experiments and Results}

\subsection{Comparison with Hill Climbing Approaches}

To further evaluate our method in comparison to a search-based method, we compare its performance against two hill climbing approaches. Therefore, we run these approaches on the dataset of 1000 generated levels and compare the performances to the models listed in Table~\ref{tab:unbalancing}. The hill climbing approaches are similar to the PCGRL narrow representation and the swap-narrow (S-narrow) representation introduced in this paper.
Since they incorporate no model to predict a position on the level to change or swap, the representations based on the wide and turtle representations are not applicable. 

Our S-narrow hill climbing implementation operates as follows: During each iteration, the method randomly selects two positions and swaps the tiles. Swaps must be carried out at each iteration since there is no model to predict this behavior.
We utilize the same reward function to evaluate the balancing (Eq.~\ref{eq:reward}) as when using PCGRL.
If the reward is negative in time step $t$, the level state is reverted to the state in $t-1$.
As in the experiments with PCGRL, levels initially matching the given balancing constraint are not considered.
The execution terminates either when the desired balancing is achieved or when the maximum of 8 changes is reached. This is the same value used in our experiments with PCGRL (cf. change percentage, Section~\ref{sec:balancing}).
%Since we consider the balancing part as a fine-tuning proceeding for already generated content, it is important to restrict the number of changes, otherwise with too many changes, a completely new level would be created. Furthermore, as of PCGRL, restricting the number of changes also ensures content diversity, as the method cannot find always the same solution using an arbitrary number of changes.
Since we consider the balancing phase as a fine-tuning stage for pre-existing content, it's crucial to restrict the number of changes. Excessive changes can lead to a total overhaul, transforming the content into an entirely new level. Moreover, adhering to the PCGRL's constraint on the number of changes also promotes content diversity, as the approach cannot consistently yield identical outcomes~\cite{khalifa_mutation_2022}.

The narrow hill climbing approach operates similarly, however, due to its ability to freely exchange tiles from the pool of available tiles, it may happen that levels remain unplayable after the execution has finished.

The results in Table~\ref{tab:hill-climbing} indicate that PCGRL using the S-wide representation balances a larger portion of levels within the prescribed limit of changes compared to both hill climbing approaches.
The S-narrow hill climbing approach achieves noteworthy balancing percentages across all $b$-values, however, it was not able to achieve a better performance than PCGRL.
In contrast, the narrow hill climbing approach shows a weak performance, still clearly lagging behind the methods using swap representations.
Furthermore, a substantial number of levels remain unplayable, with this proportion being notably higher compared to the previous PCGRL narrow approach (cf. Fig.~\ref{fig:narrow-pcgrl}).

%Table~\ref{tab:execution-time} shows that, on average, execution times are notably faster compared to the hill climbing method across all values of $b$ for PCGRL.
%Both methods exhibit their swiftest execution for $b=0.5$, with performance gradually declining towards $b=0.0$. Achieving win probabilities favoring a selected player appears to pose a greater challenge within this environment. The decreasing computation times observed for PCGRL from $b=0.0$ to $b=0.5$ also mirror the findings in Table~\ref{tab:unbalancing}, where $b=0.5$ showcases the best performance. In addition, PCGRL demonstrates greater consistency in execution times for content creation, contrasting with the hill climbing method, which displays significantly higher and more dispersed standard deviations.

%\begin{table}
%    \centering
%    \caption{Comparison on execution times with a hill climbing approach. All values are given in seconds.}
%    \begin{tabular}{c|cc|cc}
%         $b$ &  \multicolumn{2}{l|}{\textbf{PCGRL S-wide}} &   \multicolumn{2}{l}{\textbf{Hill climbing (swap)}}  \\ %\hdashline
%            &  Avg. & Std.  &  Avg. & Std.  \\ \hline
%         0.0& 10.59 & 4.37 & 45.29 &  52.81 \\
%         0.1& 10.57 & 4.29 & 40.88 &  55.89 \\
%         0.2& 10.10 & 4.23 & 26.37 &  28.56 \\
%         0.3&  9.42 & 4.25 & 24.66 &  50.80 \\
%         0.4&  8.99 & 4.06 & 26.11 &  89.96 \\
%         0.5&  8.37 & 4.07 & 11.47 &  9.68  \\
%    \end{tabular}
%    \label{tab:execution-time}
%\end{table}

\begin{table}
    \centering
    \caption{Performance comparison of PCGRL with hill climbing approaches for optimizing different balancing values $b$. All values are expressed as percentages.}
    \begin{tabular}{ccccc} \toprule
          &  \textbf{PCGRL} &  \textbf{Hill Climbing} & \multicolumn{2}{c}{\textbf{Hill Climbing}}  \\ 
          $b$  & S-wide & S-narrow & narrow & unplayable\\
         \cmidrule(lr){1-1} \cmidrule(lr){2-2} \cmidrule(lr){3-3} \cmidrule(lr){4-5} 
         0.0& \textbf{36.1} & 33.7 & 5.4  & 92.4 \\
         0.1& \textbf{35.1} & 31.9 & 10.8 & 85.8 \\
         0.2& \textbf{42.9} & 36.6 & 7.9  & 89.5 \\
         0.3& \textbf{54.6} & 39.6 & 8.9  & 88.8 \\
         0.4& \textbf{60.7} & 45.2 & 12.2 & 85.9 \\
         0.5& \textbf{68.0} & 59.6 & 14.3 & 83.1\\
         \bottomrule
    \end{tabular}
    \label{tab:hill-climbing}
\end{table}

%\begin{table}
%    \centering
%    \caption{Performance comparison of PCGRL with hill climbing approaches for optimizing different balancing values $b$. All values are expressed as percentages.}
%    \begin{tabular}{ll|cccccc}
%        &&   \multicolumn{6}{c}{$b$} \\
%         & &  0.0 & 0.1 & 0.2 & 0.3 & 0.4 & 0.5 \\  \hline
%         & Initial $=b$         & 8.6  & 7.2  & 6.4  & 11.4 & 12.0 & 15.6 \\ \hline
%        PCGRL & \multicolumn{1}{|l|}{Balanced to $b$ }    & 36.1 & 35.1 & 42.9 & 54.6 & 60.7 & 68.0 \\
%        S-wide & \multicolumn{1}{|l|}{Improved to $b$} & 71.1 & 73.5 & 76.3 & 77.9 & 81.4 & 88.9 \\
%         & \multicolumn{1}{|l|}{Baseline} & - & - & - & - & - & 48.1\\ \hline
%       Hill -  & \multicolumn{1}{|l|}{S-narrow} & 30.7 & 0.0 & 5.4 & 0.9 & 0.9 & 50.7  \\
%       Climbing  & \multicolumn{1}{|l|}{narrow}   & - & - & - & - & - & - \\
%    \end{tabular}
%    \label{tab:execution-time}
%\end{table}

\subsection{Evaluation of the Asymmetries of the Generated Levels}
\label{sec:asymmetries}
The motivation behind this work includes the creation of asymmetric game levels to enhance diversity for competitive play, which is also of general interest when implementing a PCG method~\cite{togelius_search-based_2011}. For all generated levels per model, we therefore evaluate their diversity and asymmetry with matrix comparisons. For each $b$, we determine a content diversity of 100\% of the balanced levels, which means that no level is identical to another one.

We evaluate the asymmetries across all generated levels of all models outlined in Section~\ref{sec:unbalancing} regarding their symmetric properties along the horizontal, vertical, diagonal, and counter-diagonal axes.
The diagonal symmetry of an $n \times n$ matrix $M$ can be evaluated using the Frobenius norm with $|| M - M^T||_F$ which computes the sum of the absolute differences of the matrix elements~\cite{golub2013matrix}. As we are dealing with non-numerical tile representations, we use a distance metric $h$ which returns 1 for an element $i$ if $| M_i - M^T_i | > 0$, otherwise 0. To compute symmetries along all four axes, we rotate or transpose $M$ accordingly which yields a matrix $M'$. The symmetry between $M$ and $M'$ can then be computed using Eq.~\ref{eq:symmetry}. Lower values indicate greater symmetry, a value of 0 indicates perfect symmetry.

%To overcome this limitation, we count the number of unequal tiles along a symmetry axis relative to the size of the matrix. This concept is expressed by Eq.~}\ref{eq:symmetry}\hl{, being defined in $[0,1]$. A value of 1 denotes a symmetrical level, while 0 signifies an asymmetrical one. The matrix $M'$ is derived from $M$ according to the selection of the axis for symmetry evaluation process.}

\begin{equation}
\label{eq:symmetry}
sym(M, M') = ||\; h(M - M')\; ||_F
% sym(M, M') = \left( \sum_{i=1}^{n} h(M_i, M'_i) \right) \cdot \frac{1}{n^2}
\end{equation}

% The auxiliary function $h$ returns 1 if $M_i = M'_i$, otherwise 0.
$M_{rot}$ denotes $M$ rotated counterclockwise by 90 or 180 degrees respectively, and $M^T$ the transposed matrix of $M$. To compute the symmetry score for the diagonal axis we compute $sym(M, M^T)$, for the counter-diagonal $sym(M, (M^T)_{rot180})$, the vertical $sym(M, (M_{rot90})^T)$, and the horizontal $sym(M, (M^T)_{rot90})$. The mean, standard deviation, minimal, and maximum values are presented for the four symmetry axes in Table~\ref{tab:symmetries}.

\begin{table}
    \centering
    \caption{Overview of symmetry scores for generated levels across the four axes using Eq.~\ref{eq:symmetry}. Lower values indicate more symmetry along an axis, 0 indicates perfect symmetry.}
    \begin{tabular}{ccccc} \toprule
           &  \textbf{diagonal}&  \textbf{counter-diagonal} & \textbf{vertical} & \textbf{horizontal} \\
    \cmidrule(lr){1-1} \cmidrule(lr){2-2} \cmidrule(lr){3-3} \cmidrule(lr){4-4} \cmidrule(lr){5-5}           
         Mean &  4.52 &  4.51 &  4.95 & 4.96 \\
         Std. &  0.41 &  0.40 &  0.40 & 0.39 \\
         Min. &  2.83 &  2.83 &  3.16   & 3.46   \\
         Max. &  5.48 &  5.48 &  6.0 & 6.0 \\ \bottomrule
    \end{tabular}
    \label{tab:symmetries}
\end{table}

The results show that none of the generated levels for any model is symmetric.

\section{Discussion and Limitations}
\label{sec:discussion}
% In this work, we proposed a method for automated level balancing with RL. We showed our method using swap-based representations performs better within less training steps compared to the original PCGRL.
Our experiments showed promising results within fewer training steps compared to the original PCGRL, while being more robust to ensure playability. There are, however, several things that need to be discussed.

The balancing agent's reward represents the balance state of multiple simulations on the current level state.
By design, this is done based on the information about which player has won how many times in the simulations. Incorporating game-specific information, such as player health, into the reward function could potentially improve the results. This is, however, not desirable since including domain-specific information creates dependencies on the particular game. In this case, a specific reward function must be developed for each game. Furthermore, including too much or the wrong information could bias the reward, resulting in poor performance or undesirable behavior of the model. Consequently, the model would not learn what \emph{really} affects the balance, or it might exploit unforeseen loopholes.

Splitting up the level generation and balancing process into separate units yielded much better results than doing the balancing in a single step as in the original PCGRL. 
The plain PCGRL may be able to achieve comparable results with longer training, but the architecture proposed in this work converges faster with fewer policy updates. This provides evidence that our architecture is more efficient by decomposing the problem into two simpler problems. In combination with the swapping representations, the balancing process can fully focus on balancing. We compared three adapted representations from the original PCGRL and evaluated them in terms of performance. All of them can improve the balance state, while swap-turtle gave slightly worse results than the other two. Since swap-narrow and swap-wide perform comparably, we recommend using the narrow representation when faster training and generation is required. In addition, it is independent of the grid size. If the focus is on performance, we recommend using the swap-wide representation.
%\hl{Through matrix comparisons, we determined a content diversity of 100\% of the generated and balanced levels. This is a solution at the value layer, however, not at the semantic one. There may be levels that are semantically the same e.g., if a few tiles are different, which has no influence on the game. This process cannot be easily automated. A manual examination of a random subset of 50, no semantically identical levels could be found.}

Since we have shown that our approach can improve the balance of a level, it is possible to draw conclusions about which tile types have the most impact on the balance in terms of their swap frequencies.
For this domain, the swapping of resource tiles (e.g., forest) with blocking elements (stone) had the greatest impact. Swapping players' spawn positions was not a favored action. We assume that this is due to the small number of available types of this type (only two) compared to resource tiles and therefore the model had less experience to learn these swaps appropriately. Further insights into the decision-making process of the balancing agent could be gained with methods for explainable RL \cite{eberhardinger_learning_2023}.

The balance state is evaluated by simulating the game $n$-times with heuristic player agents. Of course, this metric depends on exactly these types of players. In multiplayer online games, the pairing of players of the same skill is often ensured by matchmaking algorithms~\cite{alman_theoretical_2017} that use the Elo rating system~\cite{elo_rating_1978}.
For example, the balance would be different for players of different skill or type. This is a limitation, but also an advantage. By using different types of player agents, the level could be balanced to compensate for differences in player skill by adjusting only the level, not the players themselves. This is of great interest when balancing levels for different types of players, such as a mage and a fighter. It can be applied in e.g., role-playing games, where gear levels are an indicator of the character strength. In this way, players can use their long-farmed equipment in a competitive setting, and the game could be balanced only by the environment. Future research will extend our work in this direction. The controllable content generators~\cite{earle_learning_2021} could be an applicable solution in this context.

A challenge with the proposed architecture is the computational effort during model training. In particular, this is due to the simulation steps required for reward computation, as multiple simulations are initiated for each swap made by the agent.
RL learns throughout the training trajectories, providing an optimized approach to reduce unnecessary swaps and therefore the computational cost at inference.
We compared our method to hill climbing approaches, which rely heavily on randomness and require starting from scratch with each new iteration.
The comparison showed PCGRL's superior performance in balancing levels across various balancing values within the same dataset, while adhering to the same limit on allowed changes per episode.
The comparison between two hill climbing methods — one using the introduced swap narrow approach and the other using PCGRL's narrow representation — showed notably superior performance for the swap narrow representation. This also highlights the benefit of the introduced swapping mechanism for the fine-tuning of existing content.

In addition, we found that the reward is mostly sparse in training. This indicates that only a small subset of the action space are actions that affect the balance. Especially at the beginning of the training, the model must learn this.
To speed up the training process, we consider methods that reduce the computational effort. This would also be of great interest for the application in more complex environments.
One solution could be to reward the agent only after several time steps or even to use sparse rewards. Although the learning process would then be harder for the agent, simulation steps for the reward would be omitted. This would speed up the training process and allow the agent to explore faster.
Another approach could be to use a surrogate reward model to predict a level's balance state. This would further speed up the training, but the model's accuracy must be high enough to give appropriate rewards to ensure a correct training. Reducing the cost of reward computation thus improves the feasibility of search-based approaches. Investigating the swap-based representation pattern incorporated into e.g., the crossovers or mutations of an evolutionary algorithm is an interesting target for future research.

\subsection{Discussion on the Application of Fairness Metrics for Game Balancing}
We've automated the decision-making process of how balanced a game is, based on a function inspired by the statistical parity metric~\cite{dwork_fairness_2012}.
This metric is commonly used in the fair machine learning community e.g., to measure how fair automated decision-making systems based on machine learning are~\cite{makhlouf_applicability_2021}. With this section, we want to open the discussion on how fairness metrics can be applied to game balancing in general. At this point, it is important to note that the concept of fairness is a social and ethical concept, and not a statistical one~\cite{chouldechova_fair_2017}. Games are, however, designed to be played by humans. Therefore, social and ethical considerations should be taken into account when applying an automated balancing process, particularly in a competitive environment with multiple players.

\paragraph{An acceptable range of win rates} So far, we have only considered the balancing process for a level to be successful if the metric exactly matches the predefined balancing goal (e.g., $b=0.5$). In many cases, however, it is not possible to measure a certain value exactly by sampling. Although we have examined how often we should sample to find a balance between the number of simulation runs and minimizing the uncertainty in estimating the actual game balance, there is still a residual uncertainty. This is also a problem that often arises when collecting data about the real world, such as surveys. To address this issue, a solution is the introduction of a bias and define the optimal value not as a single value, but as a range. The \emph{Disparate Impact} metric~\cite{feldman_certifying_2015,saleiro_aequitas_2018} expresses fairness with the ratio of a predicted outcome of two groups. The fairest ratio here would be represented by a value of 1. A commonly used range for this metric to consider an outcome as fair is~$>=0.8$ as noted by Saleiro et al.~\cite{saleiro_aequitas_2018}. That being said, applied to game balancing, values in a range to the desired value can be considered as balanced. For this work, we could define a balanced game state as: $1-(w_{p1t} - w_{p2t}) \ge 0.8$. Introducing this bias would increase the range of levels considered balanced, resulting in an even better proportion of balanced levels in our method. Since this bias has a strong influence on this proportion, it should be adapted to the game domain in terms of residual uncertainty when transferred to game balancing.

\section{Conclusion and Future Work}
\label{sec:conclusion}
In this paper, we proposed a new method to balance tile-based game levels with reinforcement learning (RL).
Therefore, we introduced a novel swap-based representation family for the PCGRL framework, where the RL agent swaps the locations of tiles in a level.
Our architecture benefits from the separation of the level generation and balancing process into two subsequent processes. The latter can thus be considered as a task for fine-tuning existing content.
In this extended paper, we discussed the application of existing fairness metrics to game balancing and thus derived the reward function for the RL from the statistical parity metric, in order to automatically measure the balance state of a level.
%and include information from simulated game runs.
Furthermore, we improved the results of the conference paper on level balancing to a proportion of balanced levels of 68\% and a proportion of improvements of 88.9\%. Compared to the original PCGRL method (30.4\%) on generated levels, our approach is easier to learn and significantly improves the results.
Our RL-based method outperforms hill climbing approaches by balancing more levels within the same number of allowed changes, thus reducing the required simulation effort.

In addition, our approach is generalized, and we have shown how one can also generate levels with a specific balance in order to generate unbalanced levels that favor a specific player. This opens up new perspectives for creating levels when e.g., adults play against children or for players with different gear levels.
Finally, by analyzing which tiles are frequently swapped, it is possible to infer which tile types are most likely to affect the balance. This provides game designers with additional empirical evidence about how the game system itself actually behaves.

For future work, we plan to empirically evaluate the heuristically estimated balance using human playtests. Additionally, we are interested in balancing for different types of players, such as a mage versus an archer, by changing only the game level. Moreover, we believe that this method is applicable not only to games, but also to other domains as well. In subsequent research, we are interested in applying it to urban planning for fair infrastructure distribution.

\section*{Acknowledgments}
We thank Christoph Kern, Frederic Gerdon, Jakob Kappenberger, and Lea Cohausz for their feedback and discussions on fairness metrics. We would also like to thank Patrick Takenaka for additional feedback and comments.

\bibliographystyle{unsrt}
\bibliography{cog.bib}

\end{document}